\documentclass[10pt,twocolumn,letterpaper]{article}

\usepackage{cvpr}
\usepackage{times}
\usepackage{epsfig}
\usepackage{graphicx}
\usepackage{amsmath}
\usepackage{amssymb}

\usepackage{algorithm}
\usepackage{float}
\usepackage[noend]{algorithmic}
\DeclareMathOperator*{\argmax}{arg\,max}
\DeclareMathOperator*{\argmin}{arg\,min}
\usepackage[normalem]{ulem}
\useunder{\uline}{\ul}{}
\usepackage[pagebackref=true,breaklinks=true,letterpaper=true,colorlinks,bookmarks=false]{hyperref}

\cvprfinalcopy 

\def\cvprPaperID{****} 

\ifcvprfinal\fi
\begin{document}

\title{C2FNAS: Coarse-to-Fine Neural Architecture Search\\for 3D Medical Image Segmentation}
\author{
Qihang Yu\textsuperscript{1$\ast$}~~~~
Dong Yang\textsuperscript{2}~~~~
Holger Roth\textsuperscript{2}~~~~\\
Yutong Bai\textsuperscript{1}~~~~
Yixiao Zhang\textsuperscript{1$\ast$}~~~~
Alan L. Yuille\textsuperscript{1}~~~~
Daguang Xu\textsuperscript{2}~~~~\vspace{.3em}\\
\textsuperscript{1} The Johns Hopkins University \qquad
\textsuperscript{2} NVIDIA \\
}

\maketitle
\thispagestyle{empty}

\begin{abstract}
\label{abstract}
3D convolution neural networks (CNN) have been proved very successful in parsing organs or tumours in 3D medical images, but it remains sophisticated and time-consuming to choose or design proper 3D networks given different task contexts. Recently, Neural Architecture Search (NAS) is proposed to solve this problem by searching for the best network architecture automatically. However, the inconsistency between search stage and deployment stage often exists in NAS algorithms due to memory constraints and large search space, which could become more serious when applying NAS to some memory and time-consuming tasks, such as 3D medical image segmentation. In this paper, we propose a \textbf{coarse-to-fine neural architecture search (C2FNAS)} to automatically search a 3D segmentation network from scratch without inconsistency on network size or input size. Specifically, we divide the search procedure into two stages: 1) the coarse stage, where we search the macro-level topology of the network, $i.e.$ how each convolution module is connected to other modules; 2) the fine stage, where we search at micro-level for operations in each cell based on previous searched macro-level topology. The coarse-to-fine manner divides the search procedure into two consecutive stages and meanwhile resolves the inconsistency. We evaluate our method on 10 public datasets from Medical Segmentation Decalthon (MSD) challenge, and achieve state-of-the-art performance with the network searched using one dataset, which demonstrates the effectiveness and generalization of our searched models.\let\thefootnote\relax\footnote{$^\ast$Work done during an internship at NVIDIA.}
\end{abstract}

\vspace{-0.4cm}
\section{Introduction}
\label{Introduction}

\begin{figure}[!t]
\centering
\includegraphics[width=0.98\linewidth]{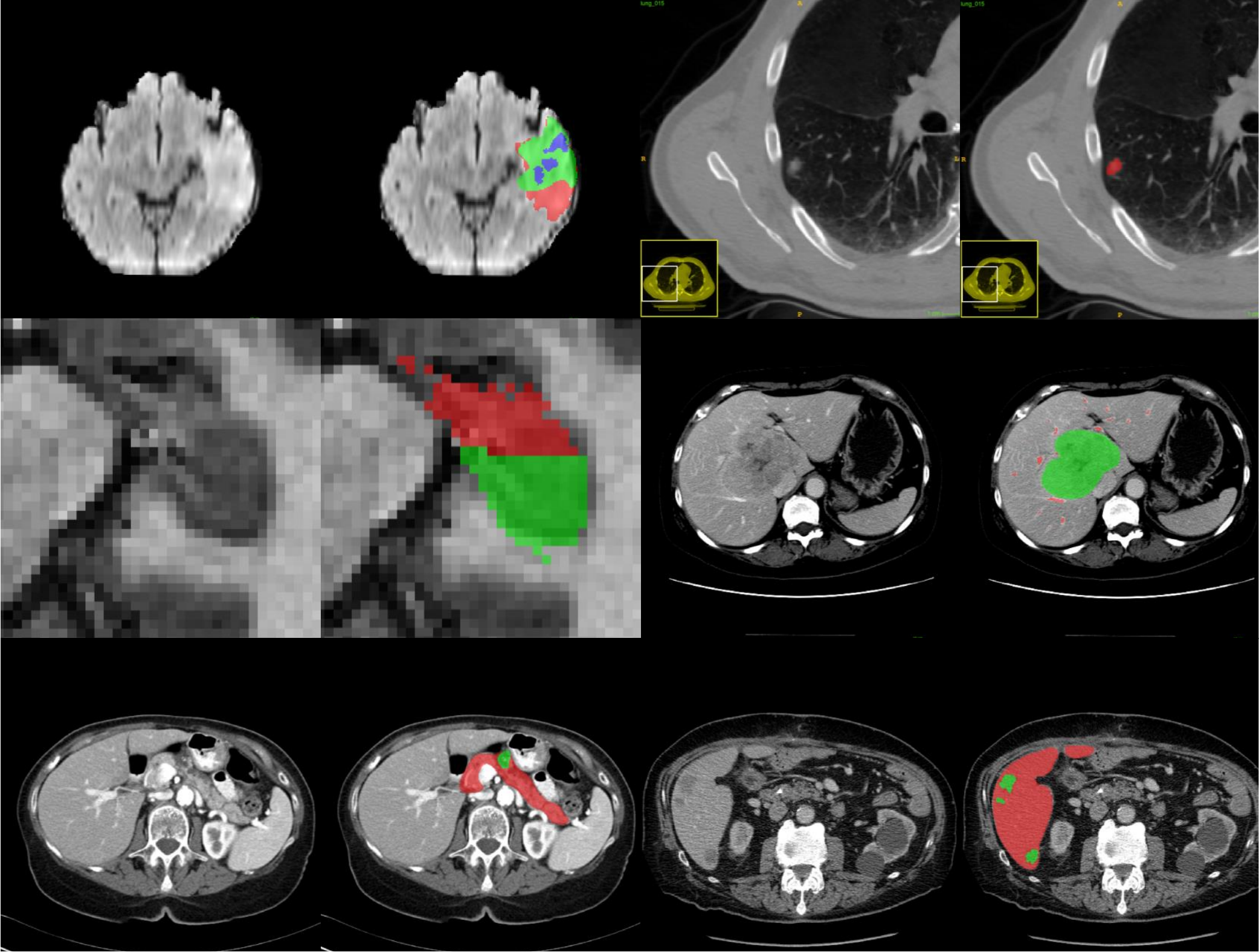}

\caption{Image and mask examples from MSD tasks (from left to right and top to bottom): brain tumours, lung tumours, hippocampus, hepatic vessel and tumours, pancreas tumours, and liver tumours, respectively. The abnormalities, texture variance, and anisotropic properties make it very challenging to achieve satisfying segmentation performance. \textcolor{red}{Red}, \textcolor{green}{green}, and \textcolor{blue}{blue} correspond to labels 1, 2 and 3, respectively, of each dataset.
\vspace{-0.4cm}
}
\label{Fig:Motivation}
\end{figure}

Medical image segmentation is an important pre-requisite of computer-aided diagnosis (CAD) which has been applied in a wide range of clinical applications. With the emerging of deep learning, great achievements have been made in this area. However, it remains very difficult to get satisfying segmentation for some challenging structures, which could be extremely small with respect to the whole volume, or vary a lot in terms of location, shape, and appearance. Besides, abnormalities, which results in a huge change in texture, and anisotropic property (different voxel spacing) make the segmentation tasks even harder. Some examples are showed in Fig~\ref{Fig:Motivation}.

\begin{figure*}[!t]
\centering
\includegraphics[width=1.0\linewidth]{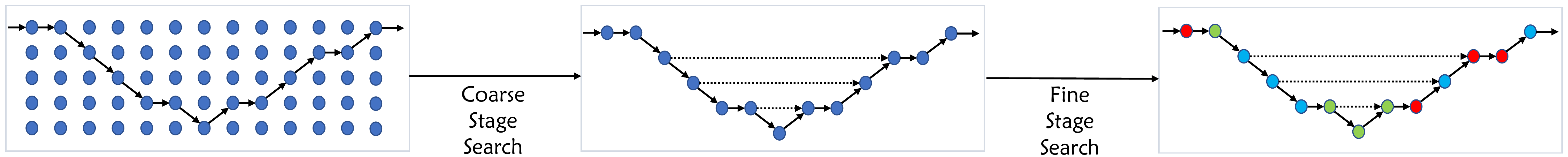}

\caption{An illustration of proposed C2FNAS. Each path from the left-most node to the right-most node is a candidate architecture. Each color represents one category of operations, $e.g.$ depthwise conv, dilated conv, or 2D/3D/P3D conv which are more common in medical image area. The dotted line indicates skip-connections from encoder to decoder. The macro-level topology is determined by coarse stage search, while the micro-level operations are further selected in fine stage search.}
\vspace{-0.2cm}
\label{Fig:Illustration}
\end{figure*}

Meanwhile, manually designing a high-performance 3D segmentation network requires adequate expertise. Most researchers are building upon existing 3D networks, such as 3D U-Net~\cite{cciccek20163d} and V-Net~\cite{milletari2016v}, with moderate modifications. In some case, an individual network is designed and only works well for certain task. To leverage this problem, Neural Architecture Search (NAS) technique is proposed in~\cite{zoph2016neural}, which aims at automatically discovering better neural network architectures than human-designed ones in terms of performance, parameters amount, or computation cost. Starting from NASNet~\cite{zoph2018learning}, many novel search spaces and search methods have been proposed~\cite{brock2017smash,ghiasi2019fpn,liu2019auto,liu2018darts,real2019regularized}. 
However, only a few works apply NAS on medical image segmentation~\cite{kim2019scalable,weng2019unet,zhu2019v}, and they only achieve a comparable performance versus those manually designed networks.

Inspired by the successful handcrafted architectures such as ResNet~\cite{he2016deep} and MobileNet~\cite{sandler2018mobilenetv2}, many NAS works focus on searching for network building blocks. However, such works usually search in a shallow network while deploying with a deeper one. An inconsistency exists in network size between the search stage and deployment stage~\cite{chen2019progressive}.~\cite{cai2018proxylessnas} and~\cite{guo2019single} avoided this problem through activating only one path at each iteration, and~\cite{chen2019progressive} proposed to progressively reduce search space and enlarge the network in order to reduce the performance gap.

Nevertheless, when the network topology is involved in the search space, things become more complex because no inconsistency is allowed in network size.~\cite{liu2019auto} incorporated the network topology into search space and relieved the memory tensity instead with a sacrifice on batch size and crop size. However, on memory-costly tasks such as 3D medical image segmentation, the memory scarcity cannot be solved by lowering the batch size or cropping size, since they are already very small compared to those of 2D tasks. Reducing them to a smaller number would lead to a much worse performance and even failure on convergence.

To avoid the inconsistency on network size or input size between search stage and deployment stage, we propose a coarse-to-fine neural architecture search scheme for 3D medical image segmentation (see Fig.~\ref{Fig:Illustration}). In detail, we divide the search procedure into coarse stage and fine stage. In the coarse stage, the search is in a small search space with limited network topologies, therefore searching in a train-from-scratch manner is affordable for each network. Moreover, 
to reduce the search space and make the search procedure more efficient, we constrain the search space under inspirations from successful medical segmentation network designs: (1) U-shape encoder-decoder structure; (2) Skip-connections between the down-sampling paths and the up-sampling paths. The search space is largely reduced with these two priors. Afterwards, we apply a topology-similarity based evolutionary algorithm considering the search space properties, which makes the searching procedure focused on the promising architecture topologies. In the fine stage, the aim is to find the best operations inside each cell. Motivated by~\cite{zhu2019v}, we let the network itself choose the operation among 2D, 3D and pseudo-3D (P3D), so that it can capture features from different viewpoints. Since the topology is already determined by coarse stage, we mitigate the memory pressure in single-path one-shot NAS manner~\cite{guo2019single}.

For validation, we apply the proposed method on ten segmentation tasks from MSD challenge~\cite{simpson2019large} and achieve state-of-the-art performance. The network is searched using the pancreas dataset which is one of the largest dataset among the 10 tasks. Our result on this proxy dataset surpasses the previous state-of-the-art by a large margin of 1\% on pancreas and 2\% on pancreas tumours. 
Then, we apply the same model and training/testing hyper-parameters across the other tasks, demonstrating the robustness and transfer-ability of the searched network.

Our contributions can be summarized into 3 folds: (1) we search a 3D segmentation network from scratch in a coarse-to-fine manner without sacrifice on network size or input size; (2) we design the specific search space and search method for each stage based on medical image segmentation priors; (3) our model achieves state-of-the-art performance on 10 datasets from MSD challenge and shows great robustness and transfer-ability.

\section{Related Work}
\label{RelatedWork}
\subsection{Medical Image Segmentation}
Deep learning based methods have achieved great success in natural image recognition~\cite{he2016deep}, detection~\cite{ren2015faster}, and segmentation~\cite{chen2018deeplab}, and they also have been dominating medical image segmentation tasks in recent years. Since U-Net was first introduced in biomedical image segmentation~\cite{ronneberger2015u}, several modifications have been proposed.~\cite{cciccek20163d} extended the 2D U-Net to a 3D version. Later, V-Net~\cite{milletari2016v} is proposed to incorporate residual blocks and soft dice loss.~\cite{oktay2018attention} introduced attention module to reinforce the U-Net model. Researchers also tried to investigate other possible architectures despite U-Net. For example, ~\cite{roth2015deeporgan,yu2018recurrent,zhou2017fixed} cut 3D volumes into 2D slices and handle them with 2D segmentation network.~\cite{liu20183d} designed a hybrid network by using ResNet50 as 2D encoder and appending 3D decoders afterwards. In~\cite{xia2018bridging}, 2D predictions are fused by a 3D network to obtain a better prediction with contextual information.

However, until now, U-Net based architectures are still the most powerful models in this area. Recently,~\cite{isensee2018nnu} introduced nnU-Net and won the first place in Medical Segmentation Decalthon (MSD) Challenge~\cite{simpson2019large}. They ensemble 2D U-Net, 3D U-Net, and cascaded 3D U-Net. The network is able to dynamically adapt itself to any given segmentation task by analysing the data attributes and adjusting hyper-parameters accordingly. 
The optimal results are achieved with different combinations of the aforementioned networks given various tasks.

\subsection{Neural Architecture Search}
Neural Architecture Search (NAS) aims at automatically discovering better neural network architectures than human-designed ones. At the beginning stage, most NAS algorithms are based on either reinforcement learning (RL)~\cite{baker2016designing,zoph2016neural,zoph2018learning} or evolutionary algorithm (EA)~\cite{real2019regularized,xie2017genetic}. In RL based methods, a controller is responsible for generating new architectures to train and evaluate, and the controller itself is trained with the architecture accuracy on validation set as reward. In EA based methods, architectures are mutated to produce better off-springs, which are also evaluated by accuracy on validation set. Since parameter sharing scheme was proposed in~\cite{pham2018efficient}, more search methods were proposed, such as differentiable NAS approaches~\cite{liu2018darts} and one-shot NAS approaches~\cite{brock2017smash}, which reduced the search cost to several GPU days or even several GPU hours.

Besides the successes NAS has achieved in natural image recognition, researchers also tried to extend it to other areas such as segmentation~\cite{liu2019auto}, detection~\cite{ghiasi2019fpn}, and attention mechanism~\cite{li2020autonl}. Moreover, there are also some works applying NAS to medical image segmentation area.~\cite{zhu2019v} designed a search space consisting of 2D, 3D, and pseudo-3D (P3D) operations, and let the network itself choose between these operations at each layer.~\cite{mortazi2018automatically,yang2019searching} use the policy gradient algorithm for automatically tuning the hyper-parameters and data augmentations. In~\cite{kim2019scalable,weng2019unet}, the cell structure is explored with a pre-defined 3D U-Net topology.

\section{Method}
\label{Method}

\subsection{Inconsistency Problem}
Early works of NAS~\cite{baker2016designing,real2019regularized,xie2017genetic,zoph2016neural,zoph2018learning} typically use a controller based on EA or RL to select network candidates from search space; then the selected architecture is trained and evaluated. Such methods need to train numerous models from scratch and thus lead to an expensive search cost. Recent works~\cite{brock2017smash,liu2018darts} propose a differentiable search method that reduces the search cost significantly, where each network is treated as a sub-network of a super-network. However, a critical problem is that the super-network cannot fit into the memory. For these methods, a trade-off is made by sacrificing the network size at search stage and building a deeper one at deployment, which results in an inconsistency problem.~\cite{cai2018proxylessnas} proposed to activate single path of the super-network at each iteration to reduce the memory cost, and~\cite{chen2019progressive} proposed to progressively increase the network size with a reduced approximate search space. However, these methods also face problems when the network topology is included in search. For instance, the progressive manner cannot deal with the network topology. As for single-path methods, since there exist illegal paths in network topology, some layers are naturally trained more times compared to others, which results in a serious fairness problem~\cite{chu2019fairnas}.

A straightforward way to solve the issue is to train each candidate from scratch respectively, yet the search cost is too expensive considering the magnitude of search space, which may contain millions of candidates or more. Auto-DeepLab~\cite{liu2019auto} introduces network topology into search space and sacrifices the input size instead of network size at training stage, where it uses a much smaller batch size and crop size. However, it introduces a new inconsistency at input size to solve the old one at network size. Besides, for memory-costly tasks such as 3D medical image segmentation, sacrificing input size is infeasible. The already small input size needs to be reduced to unreasonably smaller to fit the model in memory, which usually leads to an unstable training problem in terms of convergence, and the method only yields a random architecture finally.

\subsection{Coarse-to-fine Neural Architecture Search}
In order to resolve the inconsistency in network size and input size, and combine NAS with medical image segmentation, we develop a coarse-to-fine neural architecture search method for automatically designing 3D segmentation networks.
Without loss of generality, the architecture search space $\mathcal{A}$ consists of topology search space $\mathcal{S}$, which is represented by a directed acyclic graph (DAG), and cell operation space $\mathcal{C}$, which is represented by the color of each node in the DAG. Each network candidate is a sub-graph $s\in\mathcal{S}$ with color scheme $c\in\mathcal{C}$ and weights $w$, denoted as $\mathcal{N}(s,c,w)$.

Therefore, the search space $\mathcal{A}$ is divided into two parts: a small search space of topology $\mathcal{S}$, and a huge search space of operation $\mathcal{C}$:

\begin{equation}
    \mathcal{A} = \mathcal{S}~\times~\mathcal{C}.
\end{equation}

The topology search space is usually small and it is affordable to handle the inconsistency by training each candidate from scratch. For instance, the topology search space $\mathcal{S}$ only has up to $2.9\times 10^4$ candidates for a network with 12 cells~\cite{liu2019auto}.
The operation search space $\mathcal{C}$ can have millions of candidates, but since topology $s$ is given, techniques in NAS for recognition, $e.g.$ activating only one path at each iteration, are incorporated naturally to solve the memory limitation. 
Therefore, by regarding neural architecture search from scratch as a process of constructing a colored DAG, we divide the search procedure into two stages: (1) Coarse stage: search at macro-level for the network topology and (2) Fine stage: search for the best way to color each node, $i.e.$ finding the most suitable operation configuration.

We start with defining macro-level and micro-level. Each network consists of multiple cells, which are composed of several convolutional layers. On macro level, by defining how every cell is connected to each other, the network topology is uniquely determined. Once the topology is determined, we need to define which operation each node represents. On micro-level, we assign an operation to each node, which represents the operation inside the cell, such as standard convolution or dilated convolution.

With this two-stage procedure, we first construct a DAG representing network topology, then assign operations to each cell by coloring the corresponding node in the graph. Therefore, a network is constructed from scratch in a coarse-to-fine manner. By separating the macro-level and micro-level, we relieve the memory pressure and thus resolve the inconsistency problem between search stage and deployment stage.

\begin{figure}[!t]
\centering
\includegraphics[width=1.0\linewidth]{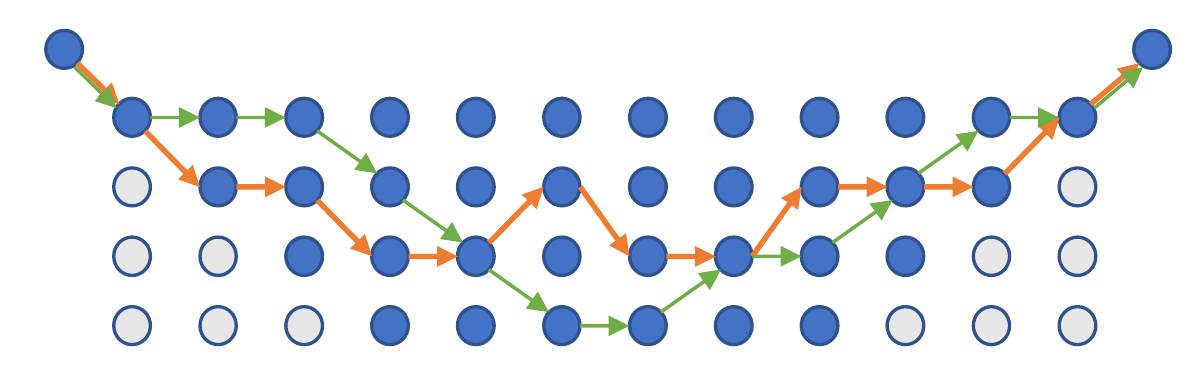}

\caption{An example of how introduced priors help reduce search space. The grey nodes are eliminated entirely from the graph. Besides, many illegal paths have been pruned off as well. An example of illegal path and legal path is shown as the orange line path and green line path separately.}
\vspace{-0.2cm}
\label{Fig:CoraseSpace}
\end{figure}

\subsection{Coarse Stage: Macro-level Search}
In this stage, we mainly focus on searching the topology of the network. A default operation is assigned to each cell, specifically standard 3D convolution in this paper, and the cell is used as the basic unit to construct the network.

Due to memory constraint and fairness problem, training a super-network and evaluating candidates with a weight-sharing method is infeasible, which means each network needs to be trained from scratch.  
The search on macro-level is formulated into a bi-level optimization with weight optimization and topology optimization:
\begin{equation}
    w_s = \argmin_{w}\mathcal{L}_{\mathrm{train}}(\mathcal{N}(s,c_0,w)),
\end{equation}
\vspace{-0.4cm}
\begin{equation}
    s^* = \argmax_{s~\in~\mathcal{S}}\mathrm{Acc_{val}}(\mathcal{N}(s,c_0,w_s)),
\end{equation}

where $s$ represents current topology and $c_0$ denotes a default coloring scheme, $e.g.$ standard 3D convolution everywhere, and $\mathcal{L}_{\mathrm{train}}$ is the loss function used at the training stage, and $\mathrm{Acc_{val}}$ the accuracy on validation set.

It is extremely time-consuming, especially considering that 3D networks have heavier computation requirements compared with 2D models. Thus, it is necessary to reduce the search space to make the search procedure more focused and efficient.

We revisit the successful medical image segmentation networks, and we find they all share something in common: (1) a U-shape encoder-decoder topology and (2) skip-connections between the down-sampling paths and the up-sampling paths. We incorporate these priors into our method and prune the search space accordingly. An illustration of how the priors help prune search space is shown in Fig.~\ref{Fig:CoraseSpace}. Therefore, the search space $\mathcal{S}$ is pruned to $\mathcal{S}'$ and the topology optimization becomes:
\begin{equation}
    \mathcal{S}' = \mathrm{PriorPrune}(\mathcal{S}),
\end{equation}
\vspace{-0.4cm}
\begin{equation}
    s^* = \argmax_{s~\in~\mathcal{S}'}\mathrm{Acc_{val}}(\mathcal{N}(s,c_0,w_s)).
\end{equation}

\begin{figure}[!t]
\centering
\includegraphics[width=1.0\linewidth]{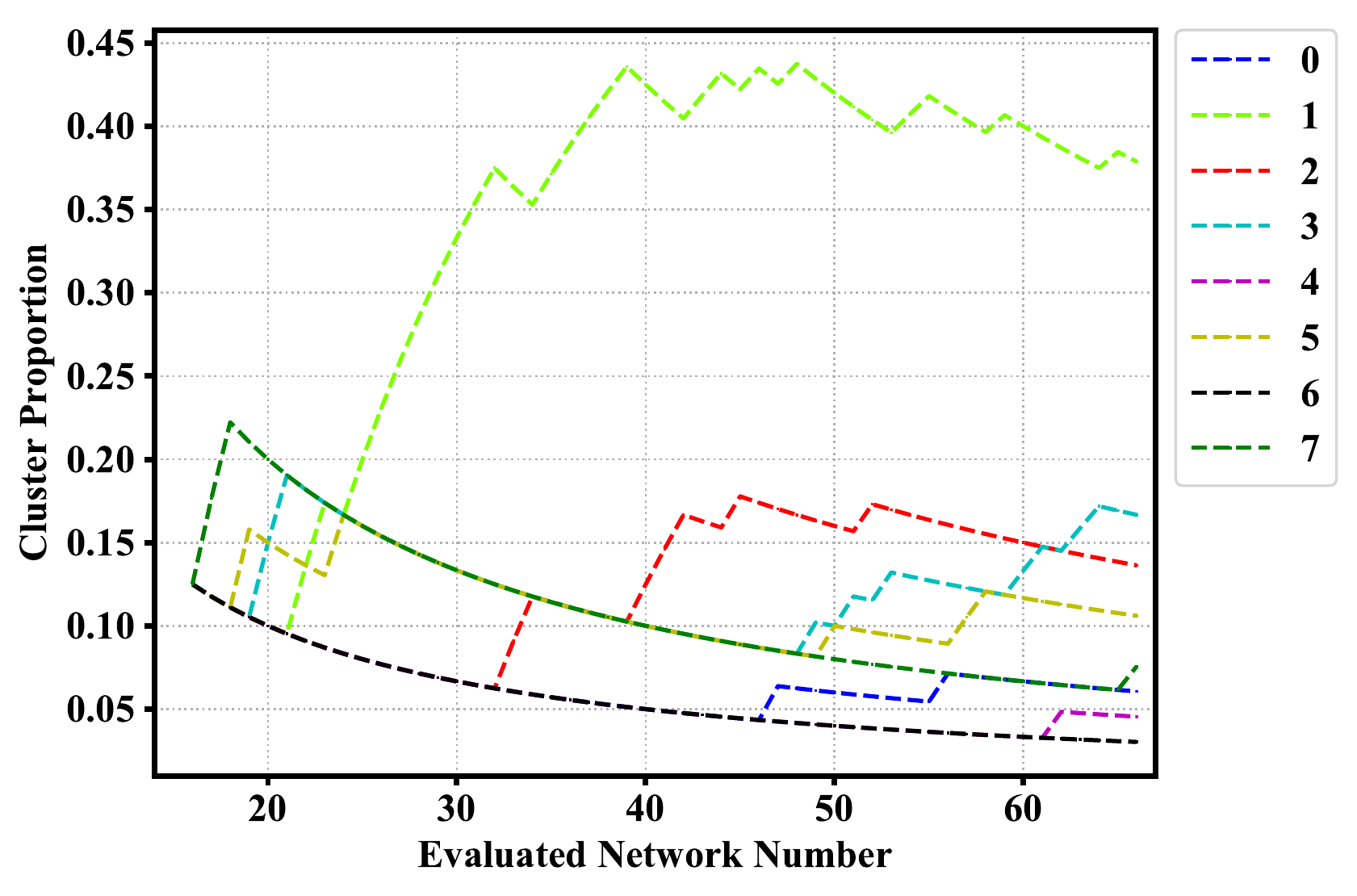}

\caption{Proportion of clusters sampled during searching at coarse stage. This figure illustrates effectiveness of the proposed evolutionary searching algorithm. Different clusters are in different colors. The x-axis label ``Evaluated Network Number" means the total number of networks trained and evaluated, while the y-axis label ``Cluster Proportion" is the proportion of number of networks belonging to a specific cluster to the total number of evaluated networks. It is shown that the algorithm gradually focuses on the most promising cluster 1, making the search procedure more efficient.} 
\vspace{-0.2cm}

\label{Fig:ClusterRatio}
\end{figure}

To further improve the search efficiency, we propose an evolutionary algorithm based on topology similarity to make use of macro-level properties. 
The idea is that with an assumption of continuous relaxation of topology search space, two similar networks should also share a similar performance. 
Specifically, we represent each network topology with a code, and we define the network similarity as the euclidean distance between two codes. Smaller the distance is, more similar two networks are. Based on the distance measurement, we classify all network candidates into several clusters with K-means algorithm~\cite{lloyd1982least} based on their encoded codes. The evolution procedure is prompted in the unit of cluster. In details, when producing next generation, we random sample some networks from each cluster, and rank the clusters by comparing performance of these networks. The higher rank a cluster is, the higher proportion of next generation will come from this cluster. As shown in Fig.~\ref{Fig:ClusterRatio}, the topology proposed by our algorithm gradually falls into the most promising cluster, demonstrating the effectiveness of it. To better make use of computation resources, we further implement this EA algorithm in an asynchronous manner as shown in Algorithm~\ref{algo1}.

\begin{algorithm}[t]
\small
\caption{Topology Similarity based Evolution}
\begin{algorithmic}[1]
\STATE $population\leftarrow$ all~topologies\\
\STATE $\mathcal{P}=\{p_1, p_2, \dots, p_k\}\leftarrow$ Cluster$(population)$ \\
\STATE history~$\mathcal{H}\leftarrow \varnothing$ \\
\STATE set of trained models $\mathcal{M} = \{m_1, m_2, \dots, m_k\}\leftarrow~\{ \varnothing \}^k$
\FOR{$i=1~to~k$}
\STATE $model.topology\leftarrow$ RandomSample$(p_i)$\\
\STATE $model.accuracy\leftarrow$ TrainEval$(model.topology)$ \\
\STATE add $model$ to $\mathcal{H}$ and $m_i$ \\
\ENDFOR
\WHILE{$|\mathcal{H}|~\leq~l$}
 \WHILE{HasIdleGPU$()$} 
 \STATE model for compare $\mathcal{D}\leftarrow \varnothing$
  \FOR{$i=1~to~k$}
  \STATE add RandomSample$(m_i)$ to $\mathcal{D}$
  \ENDFOR
 \STATE rank~$\mathcal{P}$ based on corresponding accuracy in $\mathcal{D}$\\
 \STATE $model.topology\leftarrow$ SampleUntrained$(p_{rank1})$ \\
 \STATE $model.accuracy\leftarrow$ TrainEval$(model.topology)$ \\
 \STATE add model to $\mathcal{H}$ and $m_{rank1}$ \\
 \ENDWHILE
\ENDWHILE
\RETURN highest-accuracy model in $\mathcal{H}$
\end{algorithmic}
\label{algo1}
\end{algorithm}

\subsection{Fine Stage: Micro-level Search}
After the topology of the network is determined, we further search the model at a fine-grained level by replacing the operations inside each cell. Each cell is a small fully convolutional module, which takes 1 or 2 input tensors and outputs 1 tensors. Since the topology is pre-determined in coarse stage, cell $i$ is simply represented by its operations $O_i$, which is a subset of the possible operation set $\mathcal{O}$. Our cell structure is much simpler compared with~\cite{liu2019auto}, this is because there is a trade-off between the cell complexity and cell numbers. Given the tense memory requirement of 3D models, we prefer more cells instead of a more complex cell structure.

The set of possible operations, $\mathcal{O}$, consisting of the following 3 choices: (1) $3\times3\times3$ 3D convolution; (2) $3\times3\times1$ followed by $1\times1\times3$ P3D convolution; (3) $3\times3\times1$ 2D convolution;

Considering the magnitude of fine stage search space, training each candidate from scratch is infeasible. Therefore, to address the problem of memory limitation while making search efficient, we adopt single-path one-shot NAS with uniformly sampling~\cite{guo2019single} as our search method. 
In details, we construct a super-network where each candidate is a sub-network of it, and then at each iteration of the training procedure, a candidate is uniformly sampled from the super-network and trained and updated.
After the training procedure ends, we do random search for final operation configuration. That is to say, at searching stage, we random sample $K$ candidates, and each candidate is initialized with the weights from trained super-network. All these candidates are ranked by validation performance, and the one with the highest accuracy is finally picked.

Therefore, optimization of fine stage is in single-path one-shot NAS manner with uniformly sampling, which is formulated as:
\begin{equation}
    w = \argmin_{w}\mathbb{E}_{c\in~\mathcal{C}}[\mathcal{L}_{\mathrm{train}}(\mathcal{S}(s^*,c,w))],
\end{equation}
\vspace{-0.4cm}
\begin{equation}
    c^* = \argmax_{c}\mathrm{Acc_{val}}(\mathcal{S}(s^*,c,w)),
\end{equation}
where $\mathcal{C}$ is the search space of fine stage, $i.e.$ all possibles combinations of operations.

After the coarse stage is finished, the topology $s^*$ is obtained. And the operation configuration $c^*$ comes from the fine stage. Therefore,  the final network architecture $\mathcal{N}(s^*,c^*,w)$ is constructed.

\begin{figure*}[!t]
\centering
\includegraphics[width=0.98\linewidth]{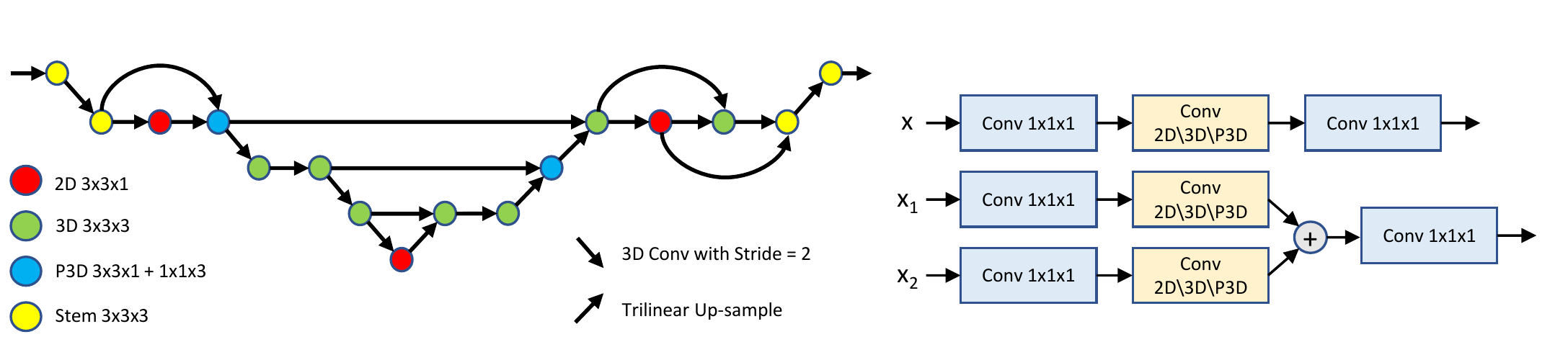}

\caption{\textbf{Left:} The final architecture of C2FNAS-Panc. \textcolor{red}{Red}, \textcolor{green}{green}, and \textcolor{blue}{blue} denote cell with 2D, 3D, P3D operations separately. \textbf{Right:} The structure of cell with single input and two inputs.}
\vspace{-0.2cm}
\label{Fig:FinalArch}
\end{figure*}

\section{Experiments}
\label{Experiments}
In this section, we firstly introduce our implementation details of C2FNAS, and then report our found architecture (searched on MSD Pancreas dataset) with semantic segmentation results on all 10 MSD datasets~\cite{simpson2019large}, which is a public comprehensive benchmark for general-purpose algorithmic validation and testing covering a large span of challenges, such as small data, unbalanced labels, large-ranging object scales, multi-class labels, and multi-modal imaging, $etc$. It contains 10 segmentation datasets, $i.e.$ Brain Tumours, Cardiac, Liver Tumours, Hippocampus, Prostate, Lung Tumours, Pancreas Tumours, Hepatic Vessels, Spleen, Colon Cancer. 

\subsection{Implementation Details}
\paragraph{Coarse Stage Search.}
At coarse stage search, the network has 12 cells at total, where 3 of them are down-sampling cells and 3 up-sampling cells, so that the model size is moderate. With the priors introduced in Section~\ref{Method}, the search space is largely reduced from $2.9\times 10^4$ to $9.24\times 10^2$.

\textbf{For network architecture}, we define one stem module at the beginning of the network, and another one at the end.
The beginning module consists of two 3D $3\times3\times3$ convolution layers, and strides are 1, 2 respectively.
The end module consists of two 3D $3\times3\times3$ convolution layers, and a trilinear up-sampling layer between the two layers. Each cell takes the output of its previous cell as input, and it will also take another input if it satisfies (1) it has a previous-previous cell at the same feature resolution level, or (2) it is the first cell after an up-sampling. In situation (1), the cell takes its previous-previous cell's output as additional input. And in situation (2), it takes the output of last cell before the corresponding down-sampling as another input, which serves as the skip-connection from encoder part to decoder part. A convolution with kernel size $1\times1\times1$ serves as pre-processing for the input. The two inputs go through convolution separately and get summed afterwards, then a $1\times1\times1$ convolution is applied to the output. The filter number starts with 32, and it is doubled after a down-sampling layer and halved after an up-sampling layer. All down-sampling operations are implemented by a $3\times3\times3$ 3D convolution with stride 2, and up-sampling by a trilinear interpolation with scale factor 2 followed by a $1\times1\times1$ convolution. Besides, in coarse stage we also set the operations in all cells to standard 3D convolution with kernel size of $3\times3\times3$.

\textbf{For evolutionary algorithm part}, we firstly represent each network topology with a code, which is a list of numbers and the length is the same as cell numbers. The number starts at 0 and increases one after a down-sampling and decreases one after an up-sampling. We use K-means algorithm to classify all candidates into 8 clusters based on the Euclidean metric of corresponding codes. At the beginning, two networks are randomly sampled from each cluster. Afterwards, whenever there is an idle GPU, one trained network is sampled from each cluster, and the cluster which the best network belongs to is picked and a new network is sampled from that cluster for training. Meanwhile, the algorithm also random samples a cluster with the probability 0.2 to add randomness and avoid local minimum. After 50 networks are evaluated, the algorithm terminates and returns the best network topology it has found.

We conduct the coarse stage search on the MSD Pancreas Tumours dataset, which contains 282 3D volumes for training and 139 for testing. The dataset is labeled with both pancreatic tumours and normal pancreas region. We divide the training data into 5 folds sequentially, where the first 4 folds for training and last fold for validation purpose. To address the anisotropic problem, we re-sample all cases to an isotropic resolution with voxel distance 1.0 $mm$ for each axis as data pre-processing.

\textbf{At training stage}, we use batch size of 8 with 8 GPUs, and patch size of $[96, 96, 96]$, where two patches are randomly cropped from each volume at each iteration. All patches are randomly rotated by $[0^\circ, 90^\circ, 180^\circ, 270^\circ]$ and flipped as data augmentation. We use SGD optimizer with learning rate of 0.02, momentum of 0.9, and weight decay of 0.00004. Besides, there is a multi-step learning rate schedule which decay the learning rate at iterations $[8000, 16000]$ with a factor 0.5. We use 1000 iterations for warm-up stage, where the learning rate increases linearly from 0.0025 to 0.02, and 20000 iterations for training. The loss function is the summation of Dice Loss and Cross-Entropy Loss, and we adopt Instance Normalization~\cite{ulyanov2016instance} and ReLU activation function. We also use Horovod~\cite{sergeev2018horovod} to speed up the multi-GPU training procedure.

\begin{table*}[!tb]
\setlength{\tabcolsep}{0.26cm}
\centering
\scriptsize
\begin{tabular}{|l||c|c|c|c|c|c|c|c|c|c|c|c|c|}
\hline
Task                  & \multicolumn{4}{c|}{Brain}          & \multicolumn{3}{c|}{Liver}      & \multicolumn{3}{c|}{Pancreas} & \multicolumn{3}{c|}{Prostate}   \\ \hline
Class                & 1              & 2              & 3              & Avg                      & 1              & 2              & Avg              & 1               & 2              & Avg              & 1              & 2              & Avg              \\ \hline
CerebriuDIKU~\cite{perslev2019one}          & \textbf{69.52} & 43.11          & 66.74          & 59.79              & 94.27          & 57.25          & 75.76          & 71.23           & 24.98          & 48.11          & 69.11          & 86.34          & 77.73          \\ \hline
Lupin                 & 66.15          & 41.63          & 64.15          & 57.31          & 94.79          & 61.40          & 78.10          & 75.99           & 21.24          & 48.62          & 72.73          & 87.62          & 80.18          \\ \hline
NVDLMED~\cite{xia20183d}               & 67.52          & 45.00          & 68.01          & 60.18         & 95.06          & 71.40          & 83.23          & 78.42           & 38.48          & 58.45          & 69.36          & 86.66          & 78.01          \\ \hline
K.A.V.athlon          & 66.63          & 46.62          & 67.46          & 60.24             & 94.74          & 61.65          & 78.20          & 74.97           & 43.20          & 59.09          & 73.42          & 87.80          & 80.61          \\ \hline
nnU-Net~\cite{isensee2018nnu}                 & 67.71          & 47.73          & 68.16          & 61.20       & \textbf{95.24} & \textbf{73.71}          & \textbf{84.48} & 79.53  & 52.27          & 65.90 & \textbf{75.81} & \textbf{89.59}          & \textbf{82.70} \\ \hline \hline
\textbf{C2FNAS-Panc}  & 67.62          & 48.56          & 69.09          & 61.76               & 94.91          & 71.63          & 83.27          & 80.59           & 52.87          & 66.73          & 73.11          & 87.43          & 80.27          \\ \hline
\textbf{C2FNAS-Panc*} & 67.62          & \textbf{48.60} & \textbf{69.72}          & \textbf{61.98}         & 94.98          & 72.89          & 83.94          & \textbf{80.76}           & \textbf{54.41}          & \textbf{67.59}          & 74.88          & 88.75          & 81.82          \\ \hline
\end{tabular}

\vspace{0.1cm}
\begin{tabular}{|l||c|c|c|c|c|c|c|c|c|c|c|c|}
\hline
Task                  & Lung                       & Heart           & \multicolumn{3}{c|}{Hippocampus}   & \multicolumn{3}{c|}{HepaticVessel} & Spleen         & Colon          & Avg (Task)      & Avg (Class)    \\ \hline
Class                & 1              & 1              & 1              & 2              & Avg              & 1                & 2              & Avg               & 1              & 1              &                &                \\ \hline
CerebriuDIKU~\cite{perslev2019one}          & 58.71          & 89.47          & 89.68          & 88.31          & 89.00          & 59.00   & 38.00          & 48.50  & 95.00 & 28.00 & 67.01          & 66.40          \\ \hline
Lupin                 & 54.61          & 91.86          & 89.66          & 88.26          & 88.96          & 60.00   & 47.00          & 53.50  & 94.00 & 9.00  & 65.61          & 65.89          \\ \hline
NVDLMED~\cite{xia20183d}                & 52.15          & 92.46          & 87.97          & 86.71          & 87.34          & 63.00            & 64.00          & 63.50           & 96.00          & 56.00          & 72.73          & 71.66          \\ \hline
K.A.V.athlon          & 60.56          & 91.72          & 89.83          & 88.52          & 89.18          & 62.00            & 63.00          &62.50           & \textbf{97.00} & 36.00          & 71.51          & 70.89          \\ \hline
nnU-Net~\cite{isensee2018nnu}                & 69.20          & \textbf{92.77}          & \textbf{90.37}          & \textbf{88.95}          & \textbf{89.66}          & 63.00            & 69.00          & 66.00           & 96.00          & 56.00          & 76.39          & 75.00          \\ \hline \hline
\textbf{C2FNAS-Panc}  & 69.47          & 92.13          & 86.87          & 85.44          & 86.16          & 63.78            & 69.41          & 66.60           & 96.60          & 55.68          & 75.87          & 74.42          \\ \hline
\textbf{C2FNAS-Panc*} & \textbf{70.44}          & 92.49 & 89.37 & 87.96          & 88.67 & \textbf{64.30}   & \textbf{71.00}          & \textbf{67.65}  & 96.28          & \textbf{58.90} & \textbf{76.97} & \textbf{75.49} \\ \hline
\end{tabular}
\caption{Comparison with state-of-the-art methods on MSD challenge test set (number from MSD leaderboard) measured by \textbf{Dice-S\o rensen coefficient (DSC)}. * denotes the 5-fold model ensemble. The numbers of tasks hepatic vessel, spleen, and colon from other teams are rounded. We also report the average on tasks and on targets respectively for an overall comparison across all tasks/targets.}
\vspace{-0.3cm}
\label{tab:MSD_Test}
\end{table*}

\textbf{At validation stage}, we test the network in a sliding window manner, where the stride = 16 for all axes. Dice-S\o rensen coefficient (DSC) metric is used to measure the performance, which is formulated as $\mathrm{DSC}$ $(\mathcal{Y},\mathcal{Z}) = \frac{2\times|\mathcal{Y}\cap \mathcal{Z}|}{|\mathcal{Y}|+|\mathcal{Z}|}$, where $\mathcal{Y}$ and $\mathcal{Z}$ denote for the prediction and ground-truth voxels set for a foreground class. The DSC has a range of $[0, 1]$ with 1 implying a perfect prediction.

\begin{table}[ht]
\centering
\footnotesize
\begin{tabular}{|l||c|c|}
\hline
Model                & Params (M) & FLOPs (G) \\ \hline
3D U-Net~\cite{cciccek20163d}             & 19.07      & 825.30    \\ \hline
V-Net~\cite{milletari2016v}                & 45.59      & 301.88    \\ \hline
VoxResNet~\cite{chen2018voxresnet}            & 6.92       & 173.02    \\ \hline
ResDSN~\cite{zhu2018a}               & 10.03      & 188.37    \\ \hline
Attention U-Net~\cite{oktay2018attention}      & 103.88     & 1162.75   \\ \hline \hline
\textbf{C2FNAS-Panc} & 17.02      & 150.78    \\ \hline
\end{tabular}
\caption{Comparison of parameters and FLOPs with other 3D networks. The FLOPs are calculated based on input size $96\times96\times96$.}
\vspace{-0.2cm}
\label{size_compare}
\end{table}

\vspace{-0.4cm}
\paragraph{Fine Stage Search.}
In the fine stage search, we mainly choose the operations from $[$2D, 3D, P3D$]$ for each cell. This search space can be large as $5.3\times 10^5$. Since the search space is numerous, we adopt a single-path one-shot NAS method based on super-network, which is trained by uniformly sampling.

The data pre-processing, data split, and training/validation setting are exactly the same as what we use in coarse stage, except that we double the number of iterations to ensure the super-network convergence. At each iteration, a random path is chosen for training. After the super-network training is finished, we random sample 2000 candidates from the search space, and use the super-network weight to initialize these candidates. Since the validation process takes a very long time due to the sliding window method, we increase the stride to 48 at all axes to speed up the search stage.

The coarse search stage takes 5 days with 64 NVIDIA V100 GPUs with 16GB memory. In fine stage, the super-network training costs 10 hours with 8 GPUs, and the searching procedure, where 2000 candidates are evaluated on validation set, takes 1 day with 8 GPUs. The large search cost is mainly because training and evaluating a 3D model itself is very time-consuming.

\vspace{-0.4cm}
\paragraph{Deployment Stage.}
The final network architecture based on the topology searched in coarse stage and operations searched in fine stage is shown in Fig.~\ref{Fig:FinalArch}. We keep the training setting same when deploying this network architecture, which means no inconsistency exists in our method. 

We use the same training setting mentioned in coarse stage, and the iteration is 40000 and multi-step decay at iterations $[16000, 32000]$. The model is trained based on same settings from scratch for each dataset, except that Prostate dataset has a very small size on Z (Axial) axis, and Hippocampus dataset has a very small shape around only 50 for each axis. Therefore we change the patch size to $128\times 128\times 32$ and stride = $[16, 16, 4]$ for Prostate, and up-sample all data to shape $96\times 96\times 96$ for Hippocampus.

\begin{table}[!tb]
\setlength{\tabcolsep}{0.28cm}
\centering
\footnotesize
\begin{tabular}{|l||c|c|c|c|}
\hline
Task          & Lung                       & \multicolumn{3}{c|}{Pancreas}                                                        \\ \hline
Class        & 1                          & 1                          & 2                          & Avg                        \\ \hline
C2FNAS-C-Lung & \textbf{71.74} & 80.26 & 52.51 & 66.39 \\ \hline
C2FNAS-C-Panc & 69.05                      & \textbf{80.39}             & 53.32                      & 66.86                      \\ \hline
C2FNAS-F-Panc & 69.77             & 80.37                      & \textbf{56.36}             & \textbf{68.37}             \\ \hline
\end{tabular}
\caption{Comparison with different stages and different proxy datasets on 5-fold cross-validation.}
\vspace{-0.3cm}
\label{tab:MSD_Val_C2F}
\end{table}

\subsection{Segmentation Results}
We report our test set results of all 10 tasks from MSD challenge and compare with other state-of-the-art methods.

Our test set results are summarized in Table~\ref{tab:MSD_Test}. We notice that other methods apply multi-model ensemble to reinforce the performance, $e.g.$ nnU-Net ensembles 5 or 10 models based on 5-fold cross-validation with one or two models, NVDLMED and CerebriuDIKU ensemble models trained from different viewpoints. Therefore, besides single-model result, we also report results with a 5-fold cross-validation model ensemble, which means 5 models are trained in a 5-fold cross-validation setting, and final test results are fused with results from these 5 models with a majority voting.

\begin{figure}[!tb]
\centering
\includegraphics[width=0.85\linewidth]{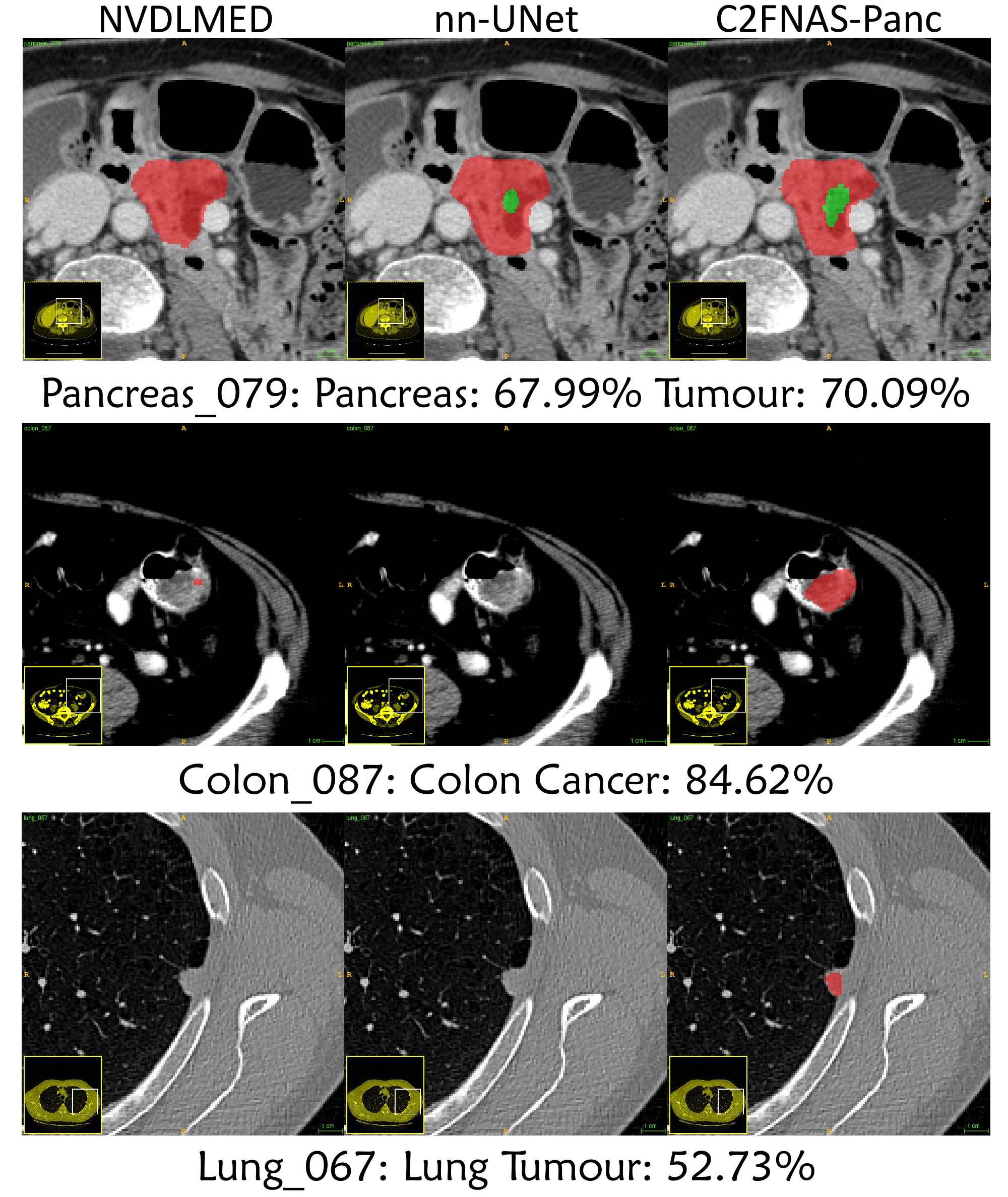}

\caption{The visualization comparison between state-of-the-art methods (1st and 2nd teams) and C2FNAS-Panc on MSD test sets. We visualize one case from each of the three most challenging tasks: pancreas and pancreas tumours, colon cancer, and lung tumours. Red denotes abnormal pancreas, colon cancer, and lung tumours respectively, and green denotes pancreas tumours. Case id and dice score of C2FNAS-Panc is at the bottom.}
\vspace{-0.2cm}
\label{Fig:Visualization}
\end{figure}

Our model shows superior performance than state-of-the-art methods on most tasks, especially the challenging ones, while enjoying a lighter model size comparing to most popular 3D models (see Table~\ref{size_compare}). We also has a higher performance in terms of average on task/class. It is noticeable that the previous state-of-the-art nnU-Net uses various kinds of data augmentation and test-time augmentation to boost the performance, while we only adopt simple data augmentation of rotation and flip, and no test-time augmentation is applied. Small datasets such as Heart and Hippocampus rely more on augmentation while a powerful architecture is easy to get over-fitting, which illustrates why our performance on these datasets does not outperform the competitors. Besides, nnU-Net uses different networks and hyper-parameters for each task, while we use the same model and hyper-parameters for all task, showing that our model is not only more powerful but also much more robust and generalizable. Some visualization comparisons are available in Fig.~\ref{Fig:Visualization}.

\section{Ablation Study}
\label{Ablation}
\subsection{Coarse Stage versus Fine Stage}
To verify the improvement of this two-stage design, we compare the performance of network from coarse stage and network from fine stage. The ``C2FNAS-C-Panc" indicates the coarse stage network searched on pancreas dataset, where the topology is searched and all operations are in standard 3D manner, while ``C2FNAS-F-Panc" is the fine stage network, where the operation configuration is searched. We compare their performance on pancreas and lung dataset with a 5-fold cross-validation. The result is shown in table~\ref{tab:MSD_Val_C2F}. It is noticeable that the fine stage search not only improves the performance on target dataset (pancreas) but also increases the model generality, thus obtains a better performance on other datasets (lung).  

\subsection{Search on Different Datasets}

Our model is searched on MSD Pancreas dataset, which contains 282 cases, and it is one of the largest dataset in MSD challenge. To verify the data number effect on our method, we also search a model topology on MSD Lung dataset, which contains 64 cases, as ablation study. The search method and hyper-parameters are same as what we use on pancreas dataset. The result is summarized in Table~\ref{tab:MSD_Val_C2F}. The ``C2FNAS-C-Lung" is the topology on lung dataset, while ``C2FNAS-C-Panc" is the topology on pancreas dataset. Topology on lung dataset performs better on lung task, while topology on pancreas dataset performs better on pancreas task. However, it is noticeable that both topologies show good performance on another dataset, demonstrating that our method works well even on a smaller dataset and the models are of great generality.

\subsection{Incorporate Model Scaling as Third Stage}
Inspired by EfficientNet~\cite{tan2019efficientnet}, we add model scaling into the search space as the third search stage. In this ablation study, we only study for scaling of filter numbers for simplicity, but a compound scaling including patch size and cell numbers is feasible. Following~\cite{tan2019efficientnet}, we adopt grid search for a channel number multiplier ranging from 0.25 to 2.0 with a step of 0.25. We report the results based on single fold validation set on pancreas and lung dataset respectively, which are summarized in Table~\ref{tab:MSD_Val_Scale}. 
It shows that model scaling can increase the model capacity and lead to a better performance. Nevertheless, scaling up the model also results in a much higher model parameters and FLOPs. Considering the large extra computation cost and to keep the model in a moderate size, we do not include model scaling into our main experiment. Yet we report it in ablation study as a potential and promising way to reinforce C2FNAS and achieve even higher performance.

\begin{table}[!tb]\
\setlength{\tabcolsep}{0.19cm}
\centering
\footnotesize
\begin{tabular}{|l||c|c|c|c|c|c|c|}
\hline
Task             & Lung           & \multicolumn{3}{c|}{Pancreas}                    & \multicolumn{3}{c|}{Hippocampus}                 \\ \hline
Class            & 1              & 1              & 2              & Avg            & 1              & 2              & Avg            \\ \hline
0.25 & 72.32          & 79.24          & 40.02          & 59.63          & 80.29          & 79.81          & 80.05          \\ \hline
0.50 & 73.89          & 80.51          & 46.34          & 63.43          & 80.74          & 80.84          & 80.79          \\ \hline
0.75 & 76.15          & 81.40          & 47.50          & 64.45          & 80.88          & 81.72          & 81.30          \\ \hline
1.00 & 74.26          & 80.74          & 49.94          & 65.34          & 81.82          & 82.10          & 81.96          \\ \hline
1.25 & 76.94          & 81.45          & 48.03          & 64.74          & 82.13          & 82.24          & 82.19          \\ \hline
1.50 & 75.37          & 81.40          & 48.87          & 65.14          & 81.02          & 81.39          & 81.21          \\ \hline
1.75 & 75.98          & 81.85          & 49.03          & 65.44          & 81.52          & 81.31          & 81.42          \\ \hline
2.00 & \textbf{77.75} & \textbf{82.18} & \textbf{50.61} & \textbf{66.40} & \textbf{82.57} & \textbf{82.34} & \textbf{82.46} \\ \hline
\end{tabular}
\caption{Influence of model scaling, the number in first column indicates the scale factor applied to model C2FNAS-Panc. The results are based on single fold of validation set and the final searched model on pancreas dataset.}
\vspace{-0.2cm}
\label{tab:MSD_Val_Scale}
\end{table}

\vspace{-0.2cm}
\section{Conclusions}
\label{Conclusions}
\vspace{-0.2cm}
In this paper, we propose to use coarse-to-fine neural architecture search to automatically design a transferable 3D segmentation network for 3D medical image segmentation, where the existing NAS methods cannot work well due to the memory-consuming property in 3D segmentation. Besides, our method, with the consistent model and hyper-parameters for all tasks, outperforms MSD champion nnU-Net, a series of well-modified and/or ensembled 2D and 3D U-Net. We do not incorporate any attention module or pyramid module, which means this is a much more powerful 3D backbone model than current popular network architectures.

{\small
\bibliographystyle{ieee_fullname}
\bibliography{egbib}
}
\clearpage

\end{document}


\section{Supplementary Material}

In this supplementary material, we provide the details of each dataset of MSD challenge. Besides, we also report the results measured in another metric Normalised Surface Distance (NSD), and a comparison of model size and computation amount with other 3D networks.

\begin{table*}[!tb]
\setlength{\tabcolsep}{0.26cm}
\centering
\small

\begin{tabular}{|l||c|c|c|c|c|c|c|c|c|c|}
\hline
Task         & \multicolumn{3}{c|}{Brain}                       & Heart          & \multicolumn{2}{c|}{Liver}      & \multicolumn{2}{c|}{Pancreas}   & \multicolumn{2}{c|}{Prostate}   \\ \hline
Class        & 1              & 2              & 3              & 1              & 1              & 2              & 1              & 2              & 1              & 2              \\ \hline
CerebriuDIKU & 88.25          & 68.98          & 88.90          & 90.63          & 96.68          & 72.60          & 91.57          & 46.43          & 94.72          & 97.90          \\ \hline
Lupin        & \textbf{88.35} & 68.06          & 89.44          & \textbf{96.84} & 98.32          & 77.41          & 94.50          & 36.72          & 94.15          & 98.24          \\ \hline
NVDLMED      & 86.99          & 69.77          & 89.82          & 95.57          & 98.26          & 87.16          & 95.22          & 57.13          & 92.96          & 97.45          \\ \hline
K.A.V.athlon & 87.80          & 72.53          & 89.50          & 94.62          & 97.93          & 79.04          & 92.49          & 65.61          & 94.77          & 98.48          \\ \hline
nnU-Net      & 87.23          & \textbf{73.31} & 90.58          & 95.90          & 98.06          & 88.40          & 95.37          & 72.78          & \textbf{95.80} & \textbf{98.90} \\ \hline \hline
C2FNAS-Panc  & 86.82          & 72.88          & 91.03          & 95.39          & 98.19          & 88.18          & 96.05          & 73.04          & 94.92          & 98.28          \\ \hline
C2FNAS-Panc* & 87.61          & 72.87          & \textbf{91.16} & 95.81          & \textbf{98.38} & \textbf{89.15} & \textbf{96.16} & \textbf{75.58} & 95.12          & 98.79          \\ \hline
\end{tabular}

\vspace{0.2cm}
\begin{tabular}{|l||c|c|c|c|c|c|c|c|c|}
\hline
Task         & Lung           & \multicolumn{2}{c|}{Hippocampus} & \multicolumn{2}{c|}{HepaticVessel} & Spleen          & Colon          & Avg (Task)     & Avg (Class)    \\ \hline
Class        & 1              & 1               & 2              & 1                & 2               & 1               & 1              &                &                \\ \hline
CerebriuDIKU & 56.10          & 97.42           & 97.42          & 38.00            & 44.00           & 98.00           & 43.00          & 77.86          & 79.51          \\ \hline
Lupin        & 55.38          & 97.72           & 97.71          & 81.00            & 54.00           & 98.00           & 16.00          & 76.31          & 78.93          \\ \hline
NVDLMED      & 50.23          & 96.07           & 96.59          & 83.00            & 72.00           & \textbf{100.00} & 66.00          & 83.19          & 84.37          \\ \hline
K.A.V.athlon & 63.95          & 97.60           & 97.47          & 83.00            & 72.00           & \textbf{100.00} & 47.00          & 82.80          & 84.34          \\ \hline
nnU-Net      & 69.13          & \textbf{97.96}  & \textbf{97.87} & 83.00            & 79.00           & 99.00           & 68.00          & 86.93          & 87.66          \\ \hline \hline
C2FNAS-Panc  & 70.53          & 95.96           & 96.37          & 83.48            & 78.90           & 98.69           & 68.95          & 86.88          & 87.51          \\ \hline
C2FNAS-Panc* & \textbf{72.22} & 97.27           & 97.35          & \textbf{83.78}   & \textbf{80.66}  & 97.66           & \textbf{72.56} & \textbf{87.83} & \textbf{88.36} \\ \hline
\end{tabular}
\caption{Comparison with state-of-the-art methods on MSD challenge test set (number from MSD leaderboard). * denotes the 5-fold model ensemble. The numbers of tasks hepatic vessel, spleen, and colon from other teams are rounded. We also report the average on tasks and on targets respectively for an overall comparison across all tasks/classes. Bigger numbers are better.}
\vspace{-0.2cm}
\label{tab:MSD_Test}
\end{table*}

\subsection{Details of Datasets}
The MSD challenge consists of 10 tasks, which can be considered as 10 separate datasets, and the proposed methods need to be effective and general enough to perform well from task to task. These tasks are:

\textbf{Task01} Brain Tumour: 484 training cases and 266 testing cases. Class 1, 2, 3 represent edema, non-enhancing tumour, enhancing tumour, respectively.

\textbf{Task02} Heart: 20 training cases and 10 testing cases. Class 1 represents left atrium.

\textbf{Task03} Liver: 131 training cases and 70 testing cases. Class 1, 2 represent liver and cancer respectively.

\textbf{Task04} Hippocampus: 260 training cases and 130 testing cases. Class 1, 2 represent anterior and posterior respectively.

\textbf{Task05} Prostate: 32 training cases and 16 testing cases. Class 1, 2 represent peripheral zone (PZ) and transition zone (TZ) respectively.

\textbf{Task06} Lung: 63 training cases and 32 testing cases. Class 1 represents cancer.

\textbf{Task07} Pancreas: 281 training cases and 139 testing cases. Class 1, 2 represent pancreas and cancer respectively.

\textbf{Task08} Hepatic Vessel: 303 training cases and 140 testing cases. Class 1, 2 represent vessel and tumour respectively.

\textbf{Task09} Spleen: 41 training cases and 20 testing cases. Class 1 represents spleen.

\textbf{Task10} Colon: 126 training cases and 64 testing cases. Class 1 represents colon cancer primaries.

It is notable that the ground truths of testing cases are not available, while test results can only be evaluated after being submitted to an evaluation server.

\subsection{Additional Results}
We also report the results measured with re-scaled Normalised Surface Distance (NSD)~\cite{simpson2019large}, which are summarized in Table~\ref{tab:MSD_Test}, the bigger number denotes a better result.

Besides, we provide a comparison about model parameters and FLOPs with other 3D segmentation networks in Table~\ref{size_compare}.

\begin{table}[]
\centering
\small
\begin{tabular}{|l||c|c|}
\hline
Model                & Params (M) & FLOPs (G) \\ \hline
3D U-Net~\cite{cciccek20163d}             & 19.07      & 825.30    \\ \hline
V-Net~\cite{milletari2016v}                & 45.59      & 301.88    \\ \hline
VoxResNet~\cite{chen2018voxresnet}            & 6.92       & 173.02    \\ \hline
ResDSN~\cite{zhu2018a}               & 10.03      & 188.37    \\ \hline
Attention U-Net~\cite{oktay2018attention}      & 103.88     & 1162.75   \\ \hline \hline
\textbf{C2FNAS-Panc} & 17.02      & 150.78    \\ \hline
\end{tabular}
\caption{Comparison of parameters and FLOPs with other 3D medical segmentation. Our model enjoys a better performance while maintaining a moderate size and computation amount compared with most 3D models. The FLOPs are calculated based on input size $96\times96\times96$.}
\label{size_compare}
\end{table}

{\small
\bibliographystyle{ieee_fullname}
\bibliography{egbib}
}